\documentclass{article}

     \usepackage[preprint,nonatbib]{neurips_2020}

\usepackage[utf8]{inputenc} 
\usepackage[T1]{fontenc}    
\usepackage{hyperref}       
\usepackage{url}            
\usepackage{booktabs}       
\usepackage{amsfonts}       
\usepackage{nicefrac}       
\usepackage{microtype}      
\usepackage[ruled,vlined,linesnumbered]{algorithm2e}
\usepackage{amsmath}
\usepackage[pdftex]{graphicx}

\title{Tricking Adversarial Attacks To Fail}

\author{%
  Blerta Lindqvist\\
  Aalto University\\
  Helsinki, Finland \\
  \texttt{blerta.lindqvist@aalto.fi} \\
}

\begin{document}

\maketitle

\begin{abstract}

Recent adversarial defense approaches have failed.
Untargeted gradient-based attacks cause classifiers to choose any wrong class.
Our novel white-box defense tricks untargeted attacks into becoming attacks targeted at designated target classes. From these target classes, we can derive the real classes. Our Target~Training defense tricks the minimization at the core of untargeted, gradient-based adversarial attacks: minimize the sum of (1) perturbation and (2) classifier adversarial loss.
Target~Training changes the classifier minimally, and trains it with additional duplicated points (at $0$ distance) labeled with designated classes. These differently-labeled duplicated samples minimize both terms (1) and (2) of the minimization, steering attack convergence to samples of designated classes, from which correct classification is derived.
Importantly, Target~Training eliminates the need to know the attack and the overhead of generating adversarial samples of attacks that minimize perturbations. We obtain an 86.2\% accuracy for CW-$L_2$($\kappa$=0) in CIFAR10, exceeding even unsecured classifier accuracy on non-adversarial samples.
Target~Training presents a fundamental change in adversarial defense strategy.

\end{abstract}


\section{Introduction}
\label{intro}

Neural network classifiers are vulnerable to malicious adversarial samples that appear indistinguishable from original samples~\cite{szegedy2013intriguing}, for example, an adversarial attack can make a traffic stop sign appear like a speed limit sign~\cite{eykholt2018robust} to a classifier. An adversarial sample created using one classifier can also fool other classifiers~\cite{szegedy2013intriguing, biggio2013evasion}, even ones with different structure and parameters~\cite{szegedy2013intriguing, goodfellow6572explaining, papernot2016transferability, tramer2017space}. This transferability of adversarial attacks~\cite{papernot2016transferability} matters because it means that classifier access is not necessary for attacks. The increasing deployment of neural network classifiers in security and safety-critical domains such as traffic~\cite{eykholt2018robust}, autonomous driving~\cite{amodei2016aisafety}, healthcare~\cite{faust2018deep}, and malware detection~\cite{cui2018detection} makes countering adversarial attacks important.

Gradient-based attacks use the classifier gradient to generate adversarial samples from non-adversarial samples. Gradient-based attacks minimize at the same time classifier adversarial loss and perturbation~\cite{szegedy2013intriguing}, though attacks can relax this minimization to allow for bigger perturbations, for example Carlini\&Wagner (CW)~\cite{carlini2017towards} for $\kappa>$0, Projected Gradient Descent (PGD)~\cite{madry2017towards}, FastGradientMethod (FGSM)~\cite{goodfellow6572explaining}. Other adversarial attacks include DeepFool~\cite{moosavi2016deepfool}, Zeroth order optimization (ZOO)~\cite{chen2017zoo}, Universal Adversarial Perturbation (UAP)~\cite{moosavi2017universal}.

Many recent proposed defenses have been broken~\cite{athalye2018obfuscated,carlini2016defensive,carlini2017adversarial,carlini2017magnet,tramer2020adaptive}. They fall largely into these categories: (1)~adversarial sample detection, (2)~gradient masking and obfuscation, (3)~ensemble, (4)~customized loss. Detection defenses~\cite{meng2017magnet,ma2018characterizing,li2018generative,hu2019new} aim to detect, correct or reject adversarial samples. Many detection defenses have been broken~\cite{carlini2017magnet,carlini2017adversarial,tramer2020adaptive}. Gradient obfuscation is aimed at preventing gradient-based attacks from access to the gradient and can be achieved by shattering gradients~\cite{guo2017countering,verma2019error,sen2020empir}, randomness~\cite{dhillon2018stochastic,li2018generative} or vanishing or exploding gradients~\cite{papernot2016distillation,song2017pixeldefend,samangouei2018defense}. Many gradient obfuscation methods have also been successfully defeated~\cite{carlini2016defensive,athalye2018obfuscated,tramer2020adaptive}. Ensemble defenses~\cite{tramer2017ensemble,verma2019error,pang2019improving,sen2020empir} have also been broken~\cite{carlini2016defensive,tramer2020adaptive}, unable to outperform their best performing component. Customized attack losses defeat defenses~\cite{tramer2020adaptive} with customized losses~\cite{pang2019rethinking,verma2019error} but also, for example ensembles~\cite{sen2020empir}. Even though it has not been defeated, Adversarial Training~\cite{szegedy2013intriguing,kurakin2016adversarial,madry2017towards} assumes that the attack is known in advance and takes time to generate adversarial samples at every iteration. The inability of recent defenses to counter adversarial attacks calls for new kinds of defensive approaches.


In this paper, we propose an adversarial defense that turns untargeted gradient-based attacks into attacks targeted at designated classes. Then our defense derives correct classification from the designated classes. Our Target~Training defense is based on the minimization~\cite{szegedy2013intriguing} at the core of untargeted gradient-based attacks. Target~Training minimizes both terms simultaneously - (1) perturbation, and (2) classifier adversarial loss - by training the classifier with nearby points that misclassify to designated classes. Thus, Target~Training guides attacks to converge to adversarial samples from designated classes. We adapt Target~Training for attacks that exclude perturbation from their minimization. Both approaches can be combined to defend against both types of attacks.

We make the following contributions:
\begin{itemize}

  \item We develop Target~Training - a novel, white-box adversarial defense that converts untargeted gradient-based attacks into attacks targeted at designated, target classes, from which correct classes are derived. Target~Training is based on the minimization at the core of untargeted gradient-based adversarial attacks.

  \item We eliminate the need to know the attack or to generate adversarial samples of a whole category of attacks. We observe that for attacks that minimize perturbation, original samples can be used instead of adversarial samples. Original samples have $0$ perturbation from themselves, the perturbation cannot be minimized further. We divide attacks into two categories: attacks that minimize perturbation; and attacks that do not.  

  \item Target Training surpasses default accuracy of 84.3\% on non-adversarial samples in CIFAR10 for most attacks that minimize perturbation. We achieve: 86.2\% for CW-$L_2$($\kappa$=0), 84.2\% for CW-$L_\infty$($\kappa$=0), 86.6\% for DeepFool, 89.0\% for ZOO and 86.8\% for UAP. For MNIST, we achieve 96.6\% for CW-$L_2$($\kappa$=0), 96.3\% for CW-$L_\infty$($\kappa$=0), 94.9\% for DeepFool, 93.0\% for ZOO and 98.6\% for UAP.
\end{itemize}

\section{Related work}
\label{rel_work}
Here, we present the state-of-the-art in adversarial attacks and defenses.

\emph{Notation} A $k$-class, neural network classifier with $\theta$ parameters is denoted by function $f(x)$ of input $x\in\mathbb{R}^{d}$ that outputs $y\in\mathbb{R}^{k}$ , where $d$ is the sample dimensionality and $k$ is number of classes. An adversarial sample is denoted by $x_{adv}$. Standard $k$-class softmax cross-entropy loss function is used to calculate the $y$ classifier output viewed as a probability distribution, each $y_i$ denoting the probability that the input belongs to class~$i$, with $0<=y_i<=1$ and $y_1+y_2+...+y_k=1$. At inference, the highest probability class predicted $C(x)=\underset{i}{\operatorname{arg\,max}} ~{y_i}$. 


\emph{Distance metrics} Adversarial attacks and defenses quantify similarity between images using norms as distance metrics, for example $L_0$ (not a real norm in the mathematical sense) - the number of pixels changed in an image, $L_2$ - the Euclidean distance, and $L_{\infty}$ - the maximum change to any pixel. 
Many attacks and defenses are not limited to any one distance metric~\cite{carlini2017towards,moosavi2016deepfool,moosavi2017universal}. 

\subsection{Adversarial attacks}

The problem of generating adversarial samples was formulated by Szegedy \emph{et al.} as a constrained minimization of the perturbation under an $L_p$ norm, such that the classification of the perturbed sample changes~\cite{szegedy2013intriguing}. Because this formulation can be hard to solve, Szegedy \emph{et al.}~\cite{szegedy2013intriguing} did a reformulation of the problem as a gradient-based, two-term minimization of the sum of the perturbation and the classifier loss:

\begin{equation} \label{eq_adv_sampl_gen}
\begin{aligned}
& {\text{minimize}}
& &  c \cdot \| x_{adv}-x\|_2^2 + loss_{f}(x_{adv},l)\\
& \text{subject to}
& &  x_{adv} \in [0,1]^n,
\end{aligned}
\end{equation}

where $l$ is an adversarial label and $c$ a constant. While term (1) ensures that the adversarial sample is visibly close to the original sample, term (2) uses the classifier gradient to minimize classifier loss on an adversarial label.

The~\ref{eq_adv_sampl_gen} (page~\pageref{eq_adv_sampl_gen}) is the foundation for gradient-based attacks, with tweaks leading to different attacks. Attack methods can use different kinds of $L_p$ norms for term (1), for example the CW attack~\cite{carlini2017towards} uses $L_0$, $L_2$ and $L_\infty$.
Some attacks do not minimize the distance from original samples, leading to adversarial samples farther from original samples. For example, the $L_\infty$ FGSM attack by GoodFellow \emph{et al.} ~\cite{goodfellow6572explaining} aims to generate adversarial samples fast and far from original samples based on an $\epsilon$ parameter that determines perturbation magnitude: $x_{adv}=x+\epsilon \cdot sign(\nabla_x loss(\theta,x,y))$.

The current strongest attack, CW~\cite{carlini2017towards}, changes the basic~\ref{eq_adv_sampl_gen} by passing the $c$ parameter to the second term and using it to tune the relative importance of the terms.
CW also introduces a confidence parameter into its minimization for the confidence of the adversarial samples. High values of confidence push CW to find adversarial samples with higher confidence that no longer minimize perturbation.
With a further change of variable, CW obtains an unconstrained minimization problem that allows it to optimize directly through back-propagation.

Implicitly following~\ref{eq_adv_sampl_gen}, Moosavi-Dezfooli \emph{et al.} define adversarial perturbation as the minimal perturbation sufficient to cause misclassification in DeepFool~\cite{moosavi2016deepfool}. DeepFool's algorithm uses the gradient to approximate linear classifier boundaries and to calculate the smallest perturbation as the smallest distance to the boundaries.
 
\emph{Black-box attacks} Black-box attacks encompass attacks that assume no access to classifier gradients. Such attacks with access to output class probabilities are called score-based attacks, for example the ZOO attack~\cite{chen2017zoo}, a black-box variant of CW~\cite{carlini2017towards}. Attacks that assume access to only the final class label decision of the classifier are called decision-based attacks, for example the Boundary~\cite{brendel2017decision} and the HopSkipJumpAttack~\cite{chen2019hopskipjumpattack} attacks.
 
\emph{Multi-step attacks} Attack perturbation can be calculated in more than one step. UAP~\cite{moosavi2017universal} finds only one universal perturbation by iterating several times over the training samples to find the minimal perturbation to move to the classifier boundary. UAP aggregates all perturbations into a universal perturbation.
The BIM attack~\cite{kurakin2016adversarial} extends FGSM~\cite{goodfellow6572explaining} by applying it iteratively with a smaller step $\alpha$.
The PGD attack~\cite{madry2017towards} is an iterative method with an $\alpha$ parameter for step-size perturbation magnitude. PGD starts at a random point $x_0$, projects the perturbation on an $L_p$-ball $B$ of a specified radius at each iteration, and clips the adversarial sample values: $x(j+1)=Proj_B(x(j)+\alpha \cdot sign(\nabla_x loss(\theta,x(j),y))$.

\subsection{Adversarial defenses}

Szegedy \emph{et al.}~used Adversarial Training~\cite{szegedy2013intriguing,kurakin2016adversarial,madry2017towards} defense to populate low probability blind spots with adversarial samples labelled correctly. Adversarial Training is one of the few non-broken defenses. Its drawback is that it needs to know the attack in advance and to train the classifier with adversarial samples of the attack.



\emph{Detection defenses} Such defenses aim to detect, then correct or reject adversarial samples. So far, adversarial samples have defied detection efforts as many detection defenses have been defeated, for example ten diverse detection methods (other network, PCA, statistical properties) by attack loss customization in~\cite{carlini2017adversarial}; attack customization against~\cite{hu2019new} by Hu \emph{et al.} in~\cite{tramer2020adaptive}; attack transferability~\cite{carlini2017magnet} against MagNet~\cite{meng2017magnet}; deep feature adversaries~\cite{sabour2015adversarial} against~\cite{roth2019odds} by Roth \emph{et al}.

\emph{Gradient masking and obfuscation} Many defenses that mask or obfuscate the classifier gradient that gradient-based attacks rely on have been defeated~\cite{carlini2016defensive,athalye2018obfuscated}. Athalye \emph{et al.}~\cite{athalye2018obfuscated} identify three types of gradient obfuscation: 
(1) Shattered gradients - incorrect gradients caused by non-differentiable components or numerical instability, for example~\cite{guo2017countering} by Guo \emph{et al.} with multiple input transformations. Athalye \emph{et al.} compute the backward pass with a function approximation that is differentiable using Backward Pass Differentiable Approximation~\cite{athalye2018obfuscated}. (2) Stochastic gradients in randomized defenses are overcome with Expectation Over Transformation~\cite{athalye2017synthesizing} by Athalye \emph{et al}. Examples of this defense are Stochastic Activation Pruning~\cite{dhillon2018stochastic} which drops layer neurons based on a weighted distribution and~\cite{xie2017mitigating} by Xie \emph{et al.} which adds a randomized layer to the input of the classifier. (3) Vanishing or exploding gradients are used, for example, in Defensive Distillation (DD)~\cite{papernot2016distillation} which reduces the amplitude of gradients of the loss function, PixelDefend~\cite{song2017pixeldefend}, Defense-GAN~\cite{samangouei2018defense}. Vanishing or exploding gradients are broken with parameters that avoid vanishing or exploding gradients~\cite{carlini2016defensive}.

\emph{Complex defenses} Defenses combining several defeated approaches, for example Li \emph{et al.}~\cite{li2018generative} using detection, randomization, multiple models and losses, can be defeated by focusing on the main defense components~\cite{tramer2020adaptive}. \cite{verma2019error,pang2019improving,sen2020empir} are defeated ensemble defenses combined with numerical instability~\cite{verma2019error} or regularization~\cite{pang2019improving} or mixed precision on weights and activations~\cite{sen2020empir}. \cite{bafna2018thwarting} uses a Fourier transform to compress inputs,~\cite{pang2019rethinking} by Pang \emph{et al.} proposes a new loss function but is defeated with a customized loss in the attack.


\textbf{Summary} Many defense approaches have been broken. They mainly focus on changing the classifier. Instead, our defense focuses on changing how attacks behave, with minimal changes to the classifier. Target~Training is the first defense that is based on the~\ref{eq_adv_sampl_gen} at the core of untargeted gradient-based adversarial attacks.

\section{Target Training}

Target Training eliminates the need to know the attack or to generate adversarial samples of attacks that minimize perturbation. Our defense turns untargeted attacks into attacks targeted at designated target classes, then derives correct classification. Target~Training undermines~\ref{eq_adv_sampl_gen} (on page~\pageref{eq_adv_sampl_gen}) of gradient-based adversarial attacks by training the classifier with exactly the points that attacks look for: nearby points (at $0$ distance) that minimize adversarial loss. For attacks that relax the minimization by removing the perturbation from it, we adjust Target~Training.

Target Training defends by training a classifier so that attacks converge to adversarial samples of designated classes. Untargeted gradient-based attacks are based on~\ref{eq_adv_sampl_gen} of the sum of (1) perturbation and (2) classifier adversarial loss. Target~Training trains the classifier with samples of designated classes that minimize both terms of the minimization at the same time, turning untargeted attacks into attacks targeted at designated target classes. From the designated classes, derivation of correct classification is straightforward.

Here, we give the intuition for categorizing adversarial attacks into: attacks that minimize perturbation; and attacks that do not minimize perturbation. For attacks that minimize perturbation, the minimization of term (1) perturbation allows Target~Training to completely eliminate the need for knowledge about the attacks or using their adversarial samples in training.
Term (1) minimization is reached at original samples because they have $0$ perturbation from themselves. In attacks that do not minimize perturbation, Target~Training needs adversarial samples during training to minimize term (1) perturbation of generated adversarial samples.

For Target Training, it is important to cause attacks to minimize both terms of~\ref{eq_adv_sampl_gen} simultaneously at samples from designated classes. Following, we outline how Target~Training minimizes term (1) perturbation for each category of attacks, and how it minimizes term (2) classifier adversarial loss. Further, we explain how Target~Training approaches for both categories of attack can be combined together.

\emph{Minimization term (1) - perturbation} Against attacks that \emph{minimize} perturbation, such as CW-$L_2$ ($\kappa=0$), CW-$L_\infty$($\kappa=0$) and DeepFool, Target~Training uses duplicates of original samples in each batch instead of adversarial samples, since no other points can have smaller distance from original samples than the original samples themselves. This removes completely the need and overhead of calculating adversarial samples against all attacks of this type. Having reduced term (1) perturbation to $0$, the minimization reduces to term (2) only. Algorithm~\ref{decoy_algo} shows classifier training against attacks that minimize perturbations.

\begin{algorithm} \label{decoy_algo} 
\SetAlgoLined
\KwResult{Target-trained classifier $N$}
 Size of the training batch is $m$, number of classes in the dataset is $k$\;
 Initialize network $N$ with double number of output classes, $2k$, keep all else in $N$ the same\;
 \Repeat{training converged}{
   Read random batch $B = \{x^1,..., x^m\}$ and its ground truth $G=\{y^1,..., y^m\}$\;   
   Duplicate batch $B$. The new batch is $B' = \{x^1,..., x^m,x^1,..., x^m\}$\;
   Duplicate the ground truth and increase ground truth values by $k$. The ground truth becomes $G'=\{y^1,..., y^m,y^1+k,..., y^m+k\}$\;
   Do one training step of network $N$ using batch $B'$\ and ground truth $G'$;
 }
 \caption{Target Training of classifier $N$ against attacks that minimize perturbation.
 }
\end{algorithm}

Against attacks that \emph{do not minimize} perturbation, such as CW-$L_2$($\kappa>0$), PGD and FGSM, Target~Training adjusts by training with additional adversarial samples from the attack. The adjusted Algorithm~\ref{target_algo_with_adv} is shown in Appendix~A.

\emph{Term (2) of the minimization - classifier adversarial loss} Let us imagine that~\ref{eq_adv_sampl_gen} of gradient-based attacks only had term (2). If this were just classifier loss without the adversarial requirement, attacks would converge to samples from the class with the highest probability, the real class. The reason is that the real class minimizes loss in a classifier that has converged. Since term (2) minimizes classifier \emph{adversarial} loss, attacks would converge to the class with the second highest probability - any of the adversarial classes with the highest probability. In a normal multi-class classifier, only the first highest probability is distinguished from the rest - a value close to $1$ for the true-label class. The rest of the classes have $\sim 0$ probability value without any distinction between them. If we could control which classes have the the top two highest probability classes, we could control the minimization of term (2).

\begin{figure}[t]
  \includegraphics[width=1.0\linewidth]{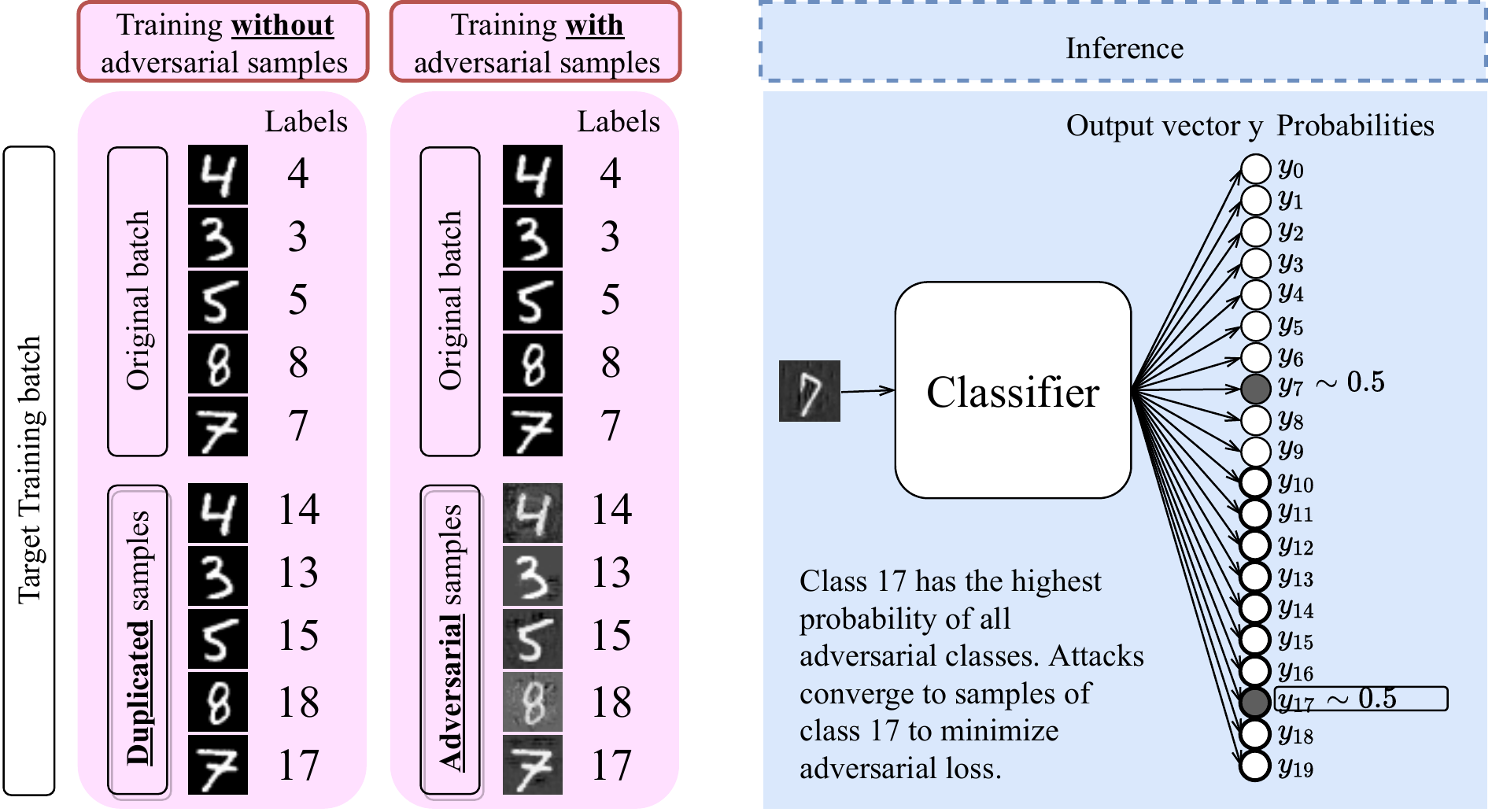}
  \caption{Target Training with and without adversarial samples, and output probabilities at inference. Example images are from the MNIST dataset, smaller batch size shown for brevity. Inference output probability values for MNIST and CIFAR10 images are shown in Appendix C, Table~\ref{target-prob-vals-mnist} and Table~\ref{target-prob-vals-cifar}.}
  \label{fig-decoy}
\end{figure}

Figure \ref{fig-decoy} shows that as a result of training with batches with additional samples that are assigned to designated target classes, the classifier has two high probability output classes at inference: original class and designated class. Since attacks minimize classifier \emph{adversarial} loss, attacks converge to adversarial samples from the designated class in order to minimize term (2) of~\ref{eq_adv_sampl_gen}. The same samples also minimize term (1) since designated classes were assigned to duplicated original samples in training. As a result, attacks converge to adversarial samples from the designated classes.

\emph{Model structure and inference} The only change to classifier structure is doubling the number of output classes from $k$ to $2k$. The loss function remains standard softmax cross entropy. Target~Training has no norm limitation because it minimizes the perturbation to $0$, which translates to $L_p$ norms of 0, for any $p$. For example, Target~Training defends against CW-$L_2$ as well as CW-$L_\infty$ attacks. Inference calculation is: $C(x)~=~\underset{i}{\operatorname{arg\,max}} ~(y_i + y_{i+k}), i \in [0 \ldots (k-1)]$.

\subsection{Simultaneous defense against both categories of attack}
\label{sect:simult}

Target~Training can be extended to counter at the same time attacks that minimize perturbations and attacks that do not. An example would be to defend against attacks that minimize perturbation, and the CW-$L_2$($\kappa=40$) attack which does not minimize perturbation. To counter both at the same time, Target~Training would triple, instead of duplicate, the batch. One set of extra samples would be original samples. The other set of extra samples would be CW-$L_2$($\kappa=40$) adversarial samples. For the labels, there would be two sets of designated classes: one set for the convergence of attacks that minimize perturbation, and the other one for the convergence of the CW-$L_2$($\kappa=40$) attack. At inference, the correct class would be: $C(x)~=~\underset{i}{\operatorname{arg\,max}} ~(y_i + y_{i+k} + y_{i+2 \cdot k}), i \in [0 \ldots (k-1)]$. This could be extended even further to accommodate more attacks that do not minimize perturbation.

\section{Experiments and results}

Our Target~Training defense leverages the fact that some attacks minimize perturbation. To counter these attacks, we replace adversarial samples with original samples because they have perturbation~$0$ from themselves. Target~Training does not use adversarial samples against attacks that minimize perturbation, but uses them against attacks that do not minimize perturbation. As a result, we conduct a separate set of experiments for each type of attack.


\emph{Threat model} We assume that the \emph{adversary goal} is to generate adversarial samples that cause untargeted misclassification. We perform white-box evaluations, assuming the adversary has complete \emph{knowledge} of the classifier and how the defense works. In terms of \emph{capabilities}, we assume that the adversary is gradient-based, has access to the CIFAR10 and MNIST image domains and is able to manipulate pixels. For attacks that minimize perturbations, no adversarial samples are used in training and no further assumption is made about attacks. For attacks that do not minimize perturbations, we assume that the attack is of the same kind as the attack used to generate the adversarial samples used during training. Further, we assume that perturbations are $L_p$-constrained.

\emph{Attack parameters} For \emph{CW}, $1,000$ iterations by default but we run experiments with up to $100,000$ iterations, confidence values are $0$ or $40$. For \emph{PGD}, we use the same attack parameters as Madry \emph{et al.} in~\cite{madry2017towards}. For MNIST, there are $40$ steps of size $0.01$, and $\epsilon=0.3$. For CIFAR10, there are $7$ steps of size $2$, and $\epsilon=8$. For \emph{ZOO} attack, we use parameters specified in the ZOO attack paper~\cite{chen2017zoo}, 1000 and 3000 iterations for CIFAR10 and MNIST, initial constant value is $0.01$, 200 adversarial samples selected randomly from the testing images of CIFAR10 and MNIST. For \emph{FGSM}, $\epsilon=0.3$, as in~\cite{madry2017towards}.

\emph{Datasets} MNIST~\cite{lecun1998mnist} and CIFAR10~\cite{krizhevsky2009cifar} are $10$-class datasets used throughout previous work. MNIST~\cite{lecun1998mnist} has $60K$, $28 \times 28 \times 1$ digit images. CIFAR10~\cite{krizhevsky2009cifar} has $70K$, $32 \times 32 \times 3$ images. All evaluations are with testing samples.

\emph{Classifier models} We purposefully do not use high capacity models, such as ResNet~\cite{he2016deep}, to show that Target~Training does not necessitate high model capacity. The architectures of MNIST and CIFAR datasets are shown in Appendix C, Table~\ref{arches}. No data augmentation used. We achieve $99.1$\% for MNIST and $84.3$\% for CIFAR10.

\emph{Tools} We generate adversarial samples with CleverHans~3.0.1~\cite{papernot2018cleverhans} for the CW~\cite{carlini2017towards}, DeepFool~\cite{moosavi2016deepfool}, and FGSM~\cite{goodfellow6572explaining} attacks and the IBM Adversarial Robustness 360 Toolbox (ART) toolbox 1.2~\cite{art2018} for the other attacks. Target~Training is written in Python~3.7.3, using Keras~2.2.4~\cite{chollet2015keras}.

\subsection{Target Training without adversarial samples against attacks that minimize perturbation}
\label{exp-no-adv}

Target Training counters adversarial attacks that minimize perturbation without using adversarial samples. The non-broken Adversarial Training defense cannot be used here because it cannot work without adversarial samples. We use an unsecured classifier as baseline because other defenses have been defeated~\cite{carlini2017magnet,carlini2017adversarial,carlini2016defensive,athalye2018obfuscated,tramer2020adaptive} successfully.

\begin{table}[t]
  \caption{Target~Training defends against attacks that minimize perturbations without using adversarial samples. In addition, Target~Training exceeds the baseline by far, and even the accuracy of unsecured classifier on non-adversarial samples in CIFAR10. Target~Training defends against attacks of different norms, against black-box attacks, and does not decrease performance for attacks with more iterations.}
  \label{decoy-no-adv-table}
  \centering
  \begin{tabular}{lrrrr}
    \toprule
                            & \multicolumn{2}{c}{CIFAR10 (84.3\%)} & \multicolumn{2}{c}{MNIST (99.1\%)} \\
                                                 \cmidrule(r){2-3}   \cmidrule(r){4-5}
                                           & \textbf{Target}   & Unsecured  & \textbf{Target}   & Unsecur. \\
    Attack                                 & \textbf{Training} & Classifier & \textbf{Training} & Classif. \\
    \midrule
    CW-$L_2$, $\kappa=0$, iterations=1K         &   85.6\% &  8.8\%     & 96.3\%   &  0.8\%    \\
    CW-$L_2$, $\kappa=0$, iterations=10K        &   86.1\% &  8.7\%     & 96.6\%   &  0.8\%    \\
    CW-$L_2$, $\kappa=0$, iterations=100K       &   86.2\% &  8.9\%     & 96.6\%   &  0.8\%    \\
    CW-$L_{\infty}$, $\kappa=0$, iterations=1K &   84.2\% & 42.0\%     & 96.3\%   & 82.1\%    \\
    DeepFool                               &   86.6\% &  9.2\%     & 94.9\%   &  1.3\%    \\
    ZOO                                    &   89.0\% & 81.5\%     & 93.0\%   & 96.0\%    \\
    UAP                                    &   86.8\% & 17.24\%    & 98.6\%   & 42.1\%    \\
    \bottomrule
  \end{tabular}
\end{table}

Table~\ref{decoy-no-adv-table} shows that Target~Training exceeds by far accuracies by unsecured classifier on adversarial samples in both CIFAR10 and MNIST. Target~Training defends against attacks that minimize perturbation without prior knowledge of such attacks and without using their adversarial samples. In CIFAR10, Target~Training exceeds even the accuracy of the unsecured classifier on non-adversarial samples (84.3\%) for most attacks. Against the ZOO black-box attack, Target~Training defense maintains its performance. Target~Training defends against attacks of different norms, for example $L_2$ and $L_{\infty}$. Finally, Target~Training improves accuracy when the attack runs more iterations. With CW-$L_2$ attack iterations from 1K-100K, accuracy increases for CIFAR10 from 85.6\% to 86.2\%, for MNIST from 96.3\% to 96.6\%.


\subsection{Target Training against adversarial attacks that do not minimize perturbation}
\label{exp-with-adv}

Against adversarial attacks that do not minimize perturbation, Target~Training uses adversarial samples and performs slightly better than Adversarial~Training. We choose Adversarial Training as a baseline because it is a non-broken adversarial defense, more details in~\nameref{rel_work}. Our implementation of Adversarial Training is based on~\cite{kurakin2016adversarial} by Kurakin \emph{et al.}, shown in Algorithm~\ref{algo_adv_training} in Appendix B.

Table~\ref{target-with-adv-table} in Appendix~C shows that Target~Training defends against attacks that do not minimize perturbation, exceeding by far accuracies of the unsecured classifier. Furthermore, Target~Training performs slightly better than Adversarial~Training against these attacks. Target~Training achieves accuracies starting from 72.1\% in CIFAR10, and 91.7\% for MNIST. In addition, Target~Training defends against multi-step attacks, in this case the PGD attack.

\subsection{Summary of results}

With our experiments in Section~\ref{exp-no-adv}, we show that we substantially improve performance against attacks that minimize perturbation without using adversarial samples.~In Section~\ref{exp-with-adv}, we show that at the same time, Target~Training maintains performance againt attacks that do not minimize perturbation, compared to previous non-broken defense. Target~Training can combine both approaches and defend simultaneously against both types of attack, as we describe in Section~\ref{sect:simult}.

\subsection{Transferability analysis}

For a defense to be strong, we need to show that it breaks the transferability of attacks~\cite{carlini2019evaluating}. A good source of adversarial samples for transferability is the unsecured classifier. We experiment on the transferability of attacks from the unsecured classifier to a classifier secured with Target~Training.

\begin{table}[t]
  \caption{Target~Training breaks the transferability of attacks from the unsecured classifier by maintaining high accuracy against attacks generated using the unsecured classifier in attacks that minimize perturbation. For attacks that do not minimize perturbation, Target~Training breaks the transferability in MNIST only.}
  \label{transf-tbl}
  \centering
  \begin{tabular}{lrrrr}
    \toprule
                         & \multicolumn{2}{c}{CIFAR10 (84.3\%)} & \multicolumn{2}{c}{MNIST (99.1\%)} \\
                                              \cmidrule(r){2-3}   \cmidrule(r){4-5}
                               & \textbf{Target}   & Unsecured  & \textbf{Target}   & Unsecured  \\
    Attack                     & \textbf{Training} & Classifier & \textbf{Training} & Classifier \\
    \midrule
    CW-$L_2$($\kappa=0$), iterations=1K       & 69.9\%  &  8.8\%   & 78.3\%   &  0.8\%     \\
    CW-$L_\infty$($\kappa=0$), iterations=1K  & 76.6\%  & 42.0\%   & 93.5\%   &  82.1\%    \\
    DeepFool                                  & 74.8\%  &  9.2\%   & 96.5\%   &  1.3\%     \\
    \midrule
    CW-$L_2$($\kappa=40$), iterations=1K      & 34.7\%  &  8.5\%   & 95.1\%   &  0.7\%     \\
    PGD                                       & 36.8\%  & 32.7\%   & 92.2\%   & 79.7\%     \\
    \bottomrule
  \end{tabular}
\end{table}

Importantly, Table~\ref{transf-tbl} shows that Target~Training breaks the transferability of adversarial samples generated by attacks that do not minimize perturbation: CW-$L_2$($\kappa=0$), CW-$L_\infty$($\kappa=0$) and DeepFool. Target~Training maintains high accuracies in CIFAR10 and MNIST against adversarial samples generated with the unsecured classifier.

Against attacks that do not minimize perturbation, CW-$L_2$($\kappa=40$) and PGD, Target~Training breaks the transferability of attacks for MNIST but not for CIFAR10. This indicates that we might need to look for samples that minimize perturbation better against this category of attacks.

\subsection{Adaptive evaluation}


Many recent defenses have failed to anticipate attacks that have defeated them~\cite{carlini2019evaluating,carlini2017adversarial,athalye2018obfuscated}. To avoid that, we perform an adaptive evaluation~\cite{carlini2019evaluating,tramer2020adaptive} of our Target~Training defense.

\emph{What attack could defeat the Target~Training defense?}~Attacks that are either targeted or not gradient-based, both outside the threat model. Most current attacks, including the strongest ones, CW and PGD, are gradient-based. Finding adversarial samples without the gradient is a hard problem~\cite{szegedy2013intriguing}.



\emph{Could Target~Training be defeated by methods used to break other defenses?}~Attack approaches~\cite{carlini2017magnet,carlini2017adversarial,carlini2016defensive,athalye2018obfuscated,tramer2020adaptive} used to defeat most current defenses cannot break Target~Training defense because we use none of the previous defenses, such as: adversarial sample detection, preprocessing, obfuscation (shattered, vanishing or exploding gradients, or randomization), ensemble, customized loss, subcomponent, non-differentiable component, or special model layers. We also keep the loss function simple - standard softmax cross-entropy and no additional loss.

\emph{Iterative attacks} decrease Target~Training accuracy more than single-step attacks, which suggests that our defense is working correctly~\cite{carlini2019evaluating}. Target~Training defends against \emph{black-box} ZOO attack, which means that we are not doing gradient masking or obfuscation~\cite{carlini2019evaluating}. Non-transferability of attacks also points to non-masking. Increasing \emph{iterations} for CW-$L_2$($\kappa=0$) 100-fold from $1K$ to $100K$ increases the defense accuracy. In CIFAR10 accuracy increases from $85.6$\% to $86.2$\%, in MNIST from $96.3$\% to $96.6$\%. This is explained by the fact that Target~Training tricks attacks into designated classes. Target~Training also maintains performance on \emph{original samples}, as shown in Appendix C, Table~\ref{target-with-adv-table-orig_smpl}. We will release the \emph{code} and trained models upon acceptance.

\section{Discussion and conclusions}

Target~Training presents a fundamental shift in adversarial defense in two ways. First, our defense is the only defense able to convert untargeted gradient-based attacks to attacks targeted at designated classes. From the designated classes, correct classification is derived. Second, Target~Training eliminates the need to know the attack in advance, and the overhead of adversarial samples, for attacks that minimize perturbation. In contrast, the previous non-broken Adversarial Training defense needs to know the attack and to generate adversarial samples of the attack during training. This is a limitation because in real applications, the attack might not be known.


Target~Training achieves high accuracy against adversarial samples and breaks the transferability of adversarial attacks. We achieve even better accuracy than 84.3\% accuracy of unsecured classifier on non-adversarial samples in CIFAR10. For example, 86.2\% for CW-$L_2$($\kappa=0$), 84.2\% for CW-$L_\infty$($\kappa=0$), 86.6\% for DeepFool, 89.0\% for ZOO and 86.8\% for UAP. We show that Target~Training breaks the transferability of adversarial samples in attacks that minimize perturbation. Target~Training also breaks the transferability of adversarial samples in attacks that do not minimize perturbation in MNIST. Target~Training also maintains performance on original, non-adversarial samples.

In conclusion, we show that Target~Training succeeds by switching the focus from changing the classifier to changing indirectly how attacks behave.





\section*{Broader impact}

Machine learning solutions in general, and neural network classifiers in particular, are increasingly being deployed into safety-critical domains, for example self-driving cars.~If attacks on such applications are possible, this impacts the safety of the systems that deploy them and the people that use them. Therefore, it is crucial to have neural network classifiers that are robust to adversarial attacks.

{\small
\bibliographystyle{ieee_fullname}
\bibliography{egbib}

\begin{thebibliography}{10}\itemsep=-1pt

\bibitem{amodei2016aisafety}
Dario Amodei, Chris Olah, Jacob Steinhardt, Paul~F. Christiano, John Schulman,
  and Dan Man{\'{e}}.
\newblock Concrete problems in {AI} safety.
\newblock {\em CoRR}, abs/1606.06565, 2016.

\bibitem{athalye2018obfuscated}
Anish Athalye, Nicholas Carlini, and David Wagner.
\newblock Obfuscated gradients give a false sense of security: Circumventing
  defenses to adversarial examples.
\newblock {\em arXiv preprint arXiv:1802.00420}, 2018.

\bibitem{athalye2017synthesizing}
Anish Athalye, Logan Engstrom, Andrew Ilyas, and Kevin Kwok.
\newblock Synthesizing robust adversarial examples.
\newblock {\em arXiv preprint arXiv:1707.07397}, 2017.

\bibitem{bafna2018thwarting}
Mitali Bafna, Jack Murtagh, and Nikhil Vyas.
\newblock Thwarting adversarial examples: An $ l\_0 $-robust sparse fourier
  transform.
\newblock In {\em Advances in Neural Information Processing Systems}, pages
  10075--10085, 2018.

\bibitem{biggio2013evasion}
Battista Biggio, Igino Corona, Davide Maiorca, Blaine Nelson, Nedim
  {\v{S}}rndi{\'c}, Pavel Laskov, Giorgio Giacinto, and Fabio Roli.
\newblock Evasion attacks against machine learning at test time.
\newblock In {\em Joint European conference on machine learning and knowledge
  discovery in databases}, pages 387--402. Springer, 2013.

\bibitem{brendel2017decision}
Wieland Brendel, Jonas Rauber, and Matthias Bethge.
\newblock Decision-based adversarial attacks: Reliable attacks against
  black-box machine learning models.
\newblock {\em arXiv preprint arXiv:1712.04248}, 2017.

\bibitem{carlini2019evaluating}
Nicholas Carlini, Anish Athalye, Nicolas Papernot, Wieland Brendel, Jonas
  Rauber, Dimitris Tsipras, Ian Goodfellow, Aleksander Madry, and Alexey
  Kurakin.
\newblock On evaluating adversarial robustness.
\newblock {\em arXiv preprint arXiv:1902.06705}, 2019.

\bibitem{carlini2016defensive}
Nicholas Carlini and David Wagner.
\newblock Defensive distillation is not robust to adversarial examples.
\newblock {\em arXiv preprint arXiv:1607.04311}, 2016.

\bibitem{carlini2017adversarial}
Nicholas Carlini and David Wagner.
\newblock Adversarial examples are not easily detected: Bypassing ten detection
  methods.
\newblock In {\em Proceedings of the 10th ACM Workshop on Artificial
  Intelligence and Security}, pages 3--14. ACM, 2017.

\bibitem{carlini2017magnet}
Nicholas Carlini and David Wagner.
\newblock Magnet and" efficient defenses against adversarial attacks" are not
  robust to adversarial examples.
\newblock {\em arXiv preprint arXiv:1711.08478}, 2017.

\bibitem{carlini2017towards}
Nicholas Carlini and David Wagner.
\newblock Towards evaluating the robustness of neural networks.
\newblock In {\em 2017 IEEE Symposium on Security and Privacy (SP)}, pages
  39--57. IEEE, 2017.

\bibitem{chen2019hopskipjumpattack}
Jianbo Chen, Michael~I Jordan, and Martin~J Wainwright.
\newblock Hopskipjumpattack: A query-efficient decision-based attack.
\newblock {\em arXiv preprint arXiv:1904.02144}, 3, 2019.

\bibitem{chen2017zoo}
Pin-Yu Chen, Huan Zhang, Yash Sharma, Jinfeng Yi, and Cho-Jui Hsieh.
\newblock Zoo: Zeroth order optimization based black-box attacks to deep neural
  networks without training substitute models.
\newblock In {\em Proceedings of the 10th ACM Workshop on Artificial
  Intelligence and Security}, pages 15--26. ACM, 2017.

\bibitem{chollet2015keras}
Fran\c{c}ois Chollet et~al.
\newblock Keras.
\newblock \url{https://keras.io}, 2015.

\bibitem{cui2018detection}
Zhihua Cui, Fei Xue, Xingjuan Cai, Yang Cao, Gai-ge Wang, and Jinjun Chen.
\newblock Detection of malicious code variants based on deep learning.
\newblock {\em IEEE Transactions on Industrial Informatics}, 14(7):3187--3196,
  2018.

\bibitem{dhillon2018stochastic}
Guneet~S Dhillon, Kamyar Azizzadenesheli, Zachary~C Lipton, Jeremy Bernstein,
  Jean Kossaifi, Aran Khanna, and Anima Anandkumar.
\newblock Stochastic activation pruning for robust adversarial defense.
\newblock In {\em International Conference on Learning Representations}, 2018.

\bibitem{eykholt2018robust}
Kevin Eykholt, Ivan Evtimov, Earlence Fernandes, Bo Li, Amir Rahmati, Chaowei
  Xiao, Atul Prakash, Tadayoshi Kohno, and Dawn Song.
\newblock Robust physical-world attacks on deep learning visual classification.
\newblock In {\em Proceedings of the IEEE Conference on Computer Vision and
  Pattern Recognition}, pages 1625--1634, 2018.

\bibitem{faust2018deep}
Oliver Faust, Yuki Hagiwara, Tan~Jen Hong, Oh~Shu Lih, and U~Rajendra Acharya.
\newblock Deep learning for healthcare applications based on physiological
  signals: a review.
\newblock {\em Computer methods and programs in biomedicine}, 2018.

\bibitem{goodfellow6572explaining}
Ian~J Goodfellow, Jonathon Shlens, and Christian Szegedy.
\newblock Explaining and harnessing adversarial examples.
\newblock {\em arXiv preprint arXiv:1412.6572}, 2014.

\bibitem{guo2017countering}
Chuan Guo, Mayank Rana, Moustapha Cisse, and Laurens Van Der~Maaten.
\newblock Countering adversarial images using input transformations.
\newblock In {\em International Conference on Learning Representations}, 2018.

\bibitem{he2016deep}
Kaiming He, Xiangyu Zhang, Shaoqing Ren, and Jian Sun.
\newblock Deep residual learning for image recognition.
\newblock In {\em Proceedings of the IEEE conference on computer vision and
  pattern recognition}, pages 770--778, 2016.

\bibitem{hu2019new}
Shengyuan Hu, Tao Yu, Chuan Guo, Wei-Lun Chao, and Kilian~Q Weinberger.
\newblock A new defense against adversarial images: Turning a weakness into a
  strength.
\newblock In {\em Advances in Neural Information Processing Systems}, pages
  1633--1644, 2019.

\bibitem{krizhevsky2009cifar}
Alex Krizhevsky, Vinod Nair, and Geoffrey Hinton.
\newblock Cifar-10 and cifar-100 datasets.
\newblock {\em URl: https://www. cs. toronto. edu/kriz/cifar. html}, 6, 2009.

\bibitem{kurakin2016adversarial}
Alexey Kurakin, Ian Goodfellow, and Samy Bengio.
\newblock Adversarial machine learning at scale.
\newblock {\em arXiv preprint arXiv:1611.01236}, 2016.

\bibitem{lecun1998mnist}
Yann LeCun, Corinna Cortes, and Christopher~JC Burges.
\newblock The mnist database of handwritten digits, 1998.
\newblock {\em URL http://yann. lecun. com/exdb/mnist}, 10:34, 1998.

\bibitem{li2018generative}
Yingzhen Li, John Bradshaw, and Yash Sharma.
\newblock Are generative classifiers more robust to adversarial attacks?
\newblock In {\em International Conference on Machine Learning}, 2019.

\bibitem{ma2018characterizing}
Xingjun Ma, Bo Li, Yisen Wang, Sarah~M Erfani, Sudanthi Wijewickrema, Grant
  Schoenebeck, Dawn Song, Michael~E Houle, and James Bailey.
\newblock Characterizing adversarial subspaces using local intrinsic
  dimensionality.
\newblock In {\em International Conference on Machine Learning}, 2018.

\bibitem{madry2017towards}
Aleksander Madry, Aleksandar Makelov, Ludwig Schmidt, Dimitris Tsipras, and
  Adrian Vladu.
\newblock Towards deep learning models resistant to adversarial attacks.
\newblock {\em arXiv preprint arXiv:1706.06083}, 2017.

\bibitem{meng2017magnet}
Dongyu Meng and Hao Chen.
\newblock Magnet: a two-pronged defense against adversarial examples.
\newblock In {\em Proceedings of the 2017 ACM SIGSAC Conference on Computer and
  Communications Security}, pages 135--147. ACM, 2017.

\bibitem{moosavi2017universal}
Seyed-Mohsen Moosavi-Dezfooli, Alhussein Fawzi, Omar Fawzi, and Pascal
  Frossard.
\newblock Universal adversarial perturbations.
\newblock In {\em Proceedings of the IEEE conference on computer vision and
  pattern recognition}, pages 1765--1773, 2017.

\bibitem{moosavi2016deepfool}
Seyed-Mohsen Moosavi-Dezfooli, Alhussein Fawzi, and Pascal Frossard.
\newblock Deepfool: a simple and accurate method to fool deep neural networks.
\newblock In {\em Proceedings of the IEEE conference on computer vision and
  pattern recognition}, pages 2574--2582, 2016.

\bibitem{art2018}
Maria-Irina Nicolae, Mathieu Sinn, Minh~Ngoc Tran, Beat Buesser, Ambrish Rawat,
  Martin Wistuba, Valentina Zantedeschi, Nathalie Baracaldo, Bryant Chen, Heiko
  Ludwig, Ian Molloy, and Ben Edwards.
\newblock Adversarial robustness toolbox v1.0.1.
\newblock {\em CoRR}, 1807.01069, 2018.

\bibitem{pang2019rethinking}
Tianyu Pang, Kun Xu, Yinpeng Dong, Chao Du, Ning Chen, and Jun Zhu.
\newblock Rethinking softmax cross-entropy loss for adversarial robustness.
\newblock In {\em International Conference on Learning Representations}, 2020.

\bibitem{pang2019improving}
Tianyu Pang, Kun Xu, Chao Du, Ning Chen, and Jun Zhu.
\newblock Improving adversarial robustness via promoting ensemble diversity.
\newblock In {\em International Conference on Learning Representations}, 2019.

\bibitem{papernot2018cleverhans}
Nicolas Papernot, Fartash Faghri, Nicholas Carlini, Ian Goodfellow, Reuben
  Feinman, Alexey Kurakin, Cihang Xie, Yash Sharma, Tom Brown, Aurko Roy,
  Alexander Matyasko, Vahid Behzadan, Karen Hambardzumyan, Zhishuai Zhang,
  Yi-Lin Juang, Zhi Li, Ryan Sheatsley, Abhibhav Garg, Jonathan Uesato, Willi
  Gierke, Yinpeng Dong, David Berthelot, Paul Hendricks, Jonas Rauber, and
  Rujun Long.
\newblock Technical report on the cleverhans v2.1.0 adversarial examples
  library.
\newblock {\em arXiv preprint arXiv:1610.00768}, 2018.

\bibitem{papernot2016distillation}
Nicolas Papernot, Patrick McDaniel, Xi Wu, Somesh Jha, and Ananthram Swami.
\newblock Distillation as a defense to adversarial perturbations against deep
  neural networks.
\newblock In {\em 2016 IEEE Symposium on Security and Privacy (SP)}, pages
  582--597. IEEE, 2016.

\bibitem{papernot2016transferability}
Nicolas Papernot, Patrick~D. McDaniel, and Ian~J. Goodfellow.
\newblock Transferability in machine learning: from phenomena to black-box
  attacks using adversarial samples.
\newblock {\em CoRR}, abs/1605.07277, 2016.

\bibitem{roth2019odds}
Kevin Roth, Yannic Kilcher, and Thomas Hofmann.
\newblock The odds are odd: A statistical test for detecting adversarial
  examples.
\newblock {\em arXiv preprint arXiv:1902.04818}, 2019.

\bibitem{sabour2015adversarial}
Sara Sabour, Yanshuai Cao, Fartash Faghri, and David~J Fleet.
\newblock Adversarial manipulation of deep representations.
\newblock In {\em International Conference on Learning Representations}, 2016.

\bibitem{samangouei2018defense}
Pouya Samangouei, Maya Kabkab, and Rama Chellappa.
\newblock Defense-gan: Protecting classifiers against adversarial attacks using
  generative models.
\newblock In {\em International Conference on Learning Representations}, 2018.

\bibitem{sen2020empir}
Sanchari Sen, Balaraman Ravindran, and Anand Raghunathan.
\newblock Empir: Ensembles of mixed precision deep networks for increased
  robustness against adversarial attacks.
\newblock In {\em International Conference on Machine Learning}, 2020.

\bibitem{song2017pixeldefend}
Yang Song, Taesup Kim, Sebastian Nowozin, Stefano Ermon, and Nate Kushman.
\newblock Pixeldefend: Leveraging generative models to understand and defend
  against adversarial examples.
\newblock In {\em International Conference on Learning Representations}, 2018.

\bibitem{szegedy2013intriguing}
Christian Szegedy, Wojciech Zaremba, Ilya Sutskever, Joan Bruna, Dumitru Erhan,
  Ian~J. Goodfellow, and Rob Fergus.
\newblock Intriguing properties of neural networks.
\newblock In {\em International Conference on Learning Representations}, 2013.

\bibitem{tramer2020adaptive}
Florian Tramer, Nicholas Carlini, Wieland Brendel, and Aleksander Madry.
\newblock On adaptive attacks to adversarial example defenses.
\newblock {\em arXiv preprint arXiv:2002.08347}, 2020.

\bibitem{tramer2017ensemble}
Florian Tram{\`e}r, Alexey Kurakin, Nicolas Papernot, Ian Goodfellow, Dan
  Boneh, and Patrick McDaniel.
\newblock Ensemble adversarial training: Attacks and defenses.
\newblock {\em arXiv preprint arXiv:1705.07204}, 2017.

\bibitem{tramer2017space}
Florian Tram{\`e}r, Nicolas Papernot, Ian Goodfellow, Dan Boneh, and Patrick
  McDaniel.
\newblock The space of transferable adversarial examples.
\newblock {\em arXiv preprint arXiv:1704.03453}, 2017.

\bibitem{verma2019error}
Gunjan Verma and Ananthram Swami.
\newblock Error correcting output codes improve probability estimation and
  adversarial robustness of deep neural networks.
\newblock In {\em Advances in Neural Information Processing Systems}, pages
  8643--8653, 2019.

\bibitem{xie2017mitigating}
Cihang Xie, Jianyu Wang, Zhishuai Zhang, Zhou Ren, and Alan Yuille.
\newblock Mitigating adversarial effects through randomization.
\newblock In {\em International Conference on Learning Representations}, 2018.

\end{thebibliography}
}



\renewcommand{\thesection}{Appendix \Alph{section}}

\setcounter{algocf}{1}
\setcounter{section}{0}

\section{Target Training algorithm against attacks that do not minimize perturbation}
\label{appx_A}

\begin{algorithm}[H] \label{target_algo_with_adv}
\SetAlgoLined
\KwResult{Target-Trained classifier $N$}
 Size of the training batch is $m$, number of classes in the dataset is $k$\;
 Initialize network $N$ with $2k$ output classes\;
 ATTACK is an adversarial attack\;
 \Repeat{training converged}{
   Read random batch $B = \{x^1,..., x^m\}$ with ground truth $G=\{y^1,..., y^m\}$  from training set\;
   Generate adversarial samples $A = ATTACK(B)$ using current state of $N$\;
   The new batch is $B' = B \bigcup A = \{x^1,..., x^m,x^1_{adv},..., x^m_{adv}\}$\;
   Duplicate the ground truth and increase the duplicate values by $k$. The ground truth becomes $G'=\{y^1,..., y^m,y^1+k,..., y^m+k\}$\;
   Do one training step of network $N$ using batch $B'$\ and ground truth $G'$;
 }
 \caption{Target Training of classifier $N$ using adversarial samples.
 }
\end{algorithm}

\section{Adversarial Training algorithm we use for comparison}
\label{appx_B}

\begin{algorithm}[H] \label{algo_adv_training}
\SetAlgoLined
\KwResult{Adversarially-Trained network $N$}
 Size of the training batch is $m$, number of classes in the dataset is $k$\;
 Initialize network $N$ with $k$ output classes\;
 \Repeat{training converged}{
   Read random batch $B = \{x^1,..., x^m\}$ with ground truth $G=\{y^1,..., y^m\}$ from training set\;
   Generate adversarial samples $\{x^1_{adv},..., x^m_{adv}\}$ from batch using current state of $N$\;
   Make new batch $B'=\{x^1,..., x^m,x^1_{adv},..., x^m_{adv}\}$\;
   Make new ground truth $G'=\{y^1,..., y^m,y^1,..., y^m\}$\;
   Do one training step of network $N$ using batch $B'$\ and ground truth $G'$;
 }
 \caption{Adversarial Training of classifier $N$ using adversarial samples.
 }
\end{algorithm}

\newpage
\section{Additional tables}
\label{appx_C}

\begin{table}[ht]
  \caption{Architectures of Target Training classifiers for CIFAR10 and MNIST datasets. For the convolutional layers, we use $L_2$ kernel regularizer. Notice that the final Dense.Softmax layers in both models have 20 output classes, twice the number of dataset classes. The default, unsecured classifiers have the same architectures, except the final layers have 10 output classes: Dense.Softmax 10.}
  \label{arches}
  \centering
  \begin{tabular}{ll}
    \toprule
    CIFAR10          & MNIST            \\ 
    \midrule
    Conv.ELU 3x3x32  & Conv.ReLU 3x3x32 \\ 
    BatchNorm        & BatchNorm        \\ 
    Conv.ELU 3x3x32  & Conv.ReLU 3x3x64 \\ 
    BatchNorm        & BatchNorm        \\ 
    MaxPool 2x2      & MaxPool 2x2      \\ 
    Dropout 0.2      & Dropout 0.25     \\ 
    Conv.ELU 3x3x64  & Dense 128        \\ 
    BatchNorm        & Dropout 0.5      \\ 
    Conv.ELU 3x3x64  & Dense.Softmax \textbf{20} \\ 
    BatchNorm        &                  \\ 
    MaxPool 2x2      &                  \\ 
    Dropout 0.3      &                  \\ 
    Conv.ELU 3x3x128 &                  \\ 
    BatchNorm        &                  \\ 
    Conv.ELU 3x3x128 &                  \\ 
    BatchNorm        &                  \\
    MaxPool 2x2      &                  \\
    Dropout 0.4      &                  \\ 
    Dense.Softmax \textbf{20} &                  \\
    \bottomrule
  \end{tabular}
\end{table}


\begin{table}[ht]
  \caption{Comparing Target Training and Adversarial Training accuracy on original samples. Adversarial~Training is not applicable (NA) in the first row because it needs adversarial samples.}
  \label{target-with-adv-table-orig_smpl}
  \centering
  \begin{tabular}{lrrrrrr}
    \toprule
                                & \multicolumn{3}{c}{CIFAR10 (84.3\%)} & \multicolumn{3}{c}{MNIST (99.1)} \\
                                                     \cmidrule(r){2-4}   \cmidrule(r){5-7}  
                                        & Target & Advers.& No     & Target & Advers.& No     \\
                                        & Train- & Train- & Defe-  & Train- & Train- & Defe-  \\
    Adv. samples in training            & ing    & ing    & nse    & ing    & ing    & nse    \\
    \midrule
    none (against attacks w/o perturb.) & 86.7\% &     NA & 84.3\% & 98.6\% &     NA & 84.3\% \\
    CW-$L_2$ ($\kappa=40$)              & 77.7\% & 77.4\% & 84.3\% & 98.0\% & 98.0\% & 99.1\% \\
    PGD                                 & 76.3\% & 76.9\% & 84.3\% & 98.3\% & 98.4\% & 99.1\% \\
    FGSM($\epsilon=0.3$)                & 77.6\% & 76.6\% & 84.3\% & 98.6\% & 98.6\% & 99.1\% \\
    \bottomrule
  \end{tabular}
\end{table}



\begin{table}[t]
  \caption{Class output probabilities for Target~Training on original, and adversarial samples from MNIST. Adversarial samples generated with CW-$L_2$($\kappa=0$). Zero probability values and probability values rounded to zero have been omitted.}
  \label{target-prob-vals-mnist}
  \centering
  \begin{tabular}{lrrrrrrrrrr}
    \toprule 
    &  \multicolumn{10}{c}{Original images}\\
    Labels          & \includegraphics[width=0.06\linewidth]{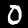} 
                    & \includegraphics[width=0.06\linewidth]{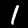} 
                    & \includegraphics[width=0.06\linewidth]{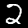} 
                    & \includegraphics[width=0.06\linewidth]{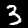} 
                    & \includegraphics[width=0.06\linewidth]{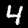}
                    & \includegraphics[width=0.06\linewidth]{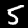}
                    & \includegraphics[width=0.06\linewidth]{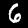} 
                    & \includegraphics[width=0.06\linewidth]{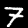}
                    & \includegraphics[width=0.06\linewidth]{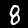} 
                    & \includegraphics[width=0.06\linewidth]{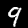} \\ 
     \midrule
     0 & 0.508 &       &       &  &  &  &  &  &  & \\
     1 &       & 0.435 &       &  &  &  &  &  &  & \\
     2 &       &       & 0.616 &  &  &  &  &  &  & \\
     3 &       &       &       & 0.776 &  &  &  &  &  & \\
     4 &       &       &       &       & 0.754 &  &  &  &  & \\
     5 &       &       &       &       &       & 0.622 &  &  &  & \\
     6 &       &       &       &       &       &       & 0.652 &  &  & \\
     7 &       &       &       &       &       &       &       & 0.614 &  & \\
     8 &       &       &       &       &       &       &       &       & 0.524 & \\
     9 &       &       &       &       &       &       &       &       &       & 0.430 \\
    10 & 0.492 &       &       &  &  &  &  &  &  & \\
    11 &       & 0.565 &       &  &  &  &  &  &  & \\
    12 &       &       & 0.384 &  &  &  &  &  &  & \\
    13 &       &       &       & 0.224 &  &  &  &  &  & \\
    14 &       &       &       &       & 0.246 &  &  &  &  & \\
    15 &       &       &       &       &       & 0.378 &  &  &  & \\
    16 &       &       &       &       &       &       & 0.348 &  &  & \\
    17 &       &       &       &       &       &       &       & 0.386 &  & \\
    18 &       &       &       &       &       &       &       &       & 0.476 & \\
    19 &       &       &       &       &       &       &       &       &       & 0.570 \\
    \midrule
    &  \multicolumn{10}{c}{Adversarial images}\\
   Labels           & \includegraphics[width=0.06\linewidth]{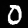} 
                    & \includegraphics[width=0.06\linewidth]{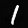} 
                    & \includegraphics[width=0.06\linewidth]{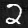} 
                    & \includegraphics[width=0.06\linewidth]{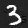} 
                    & \includegraphics[width=0.06\linewidth]{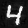}
                    & \includegraphics[width=0.06\linewidth]{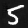}
                    & \includegraphics[width=0.06\linewidth]{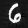} 
                    & \includegraphics[width=0.06\linewidth]{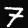}
                    & \includegraphics[width=0.06\linewidth]{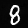} 
                    & \includegraphics[width=0.06\linewidth]{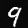} \\
     0 & 0.500 &       &       &  &  &  &  &  &  & \\
     1 &       & 0.503 &       &  &  &  &  &  &  & \\
     2 &       &       & 0.493 &  &  &  &  &  &  & \\
     3 &       &       &       & 0.492 &  &  &  &  &  & \\
     4 &       &       &       &       & 0.500 &  &  &  &  & \\
     5 &       &       &       &       &       & 0.500 &  &  &  & \\
     6 &       &       &       &       &       &       & 0.499 &  &  & \\
     7 &       &       &       &       &       &       &       & 0.457 &  & \\
     8 &       &       &       &       &       &       &       &       & 0.500 & \\
     9 &       &       &       &       &       &       &       &       &       & 0.505 \\
   10  & 0.500 &       &       &  &  &  &  &  &  & \\
   11  &       & 0.497 &       &  &  &  &  &  &  & \\
   12  &       &       & 0.507 &  &  &  &  &  &  & \\
   13  &       &       &       & 0.508 &  &  &  &  &  & \\
   14  &       &       &       &       & 0.500 &  &  &  &  & \\
   15  &       &       &       &       &       & 0.500 &  &  &  & \\
   16  &       &       &       &       &       &       & 0.501 &  &  & \\
   17  &       &       &       &       &       &       &       & 0.543 &  & \\
   18  &       &       &       &       &       &       &       &       & 0.500 & \\
   19  &       &       &       &       &       &       &       &       &       & 0.495 \\

    \bottomrule
  \end{tabular}
\end{table}


\begin{table}[t]
  \caption{Class output probabilities for Target~Training on original, and adversarial samples from CIFAR10. Adversarial samples generated with CW-$L_2$($\kappa=0$). Zero probability values and probability values rounded to zero have been omitted. The two highest class probabilities for each image are made bold. The deer (fifth image) appears to be misclassified as a horse.}    
  \label{target-prob-vals-cifar}
  \centering
  \begin{tabular}{lrrrrrrrrrr}
    \toprule 
    &  \multicolumn{10}{c}{Original images}\\
    Labels          & \includegraphics[width=0.06\linewidth]{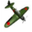} 
                    & \includegraphics[width=0.06\linewidth]{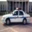} 
                    & \includegraphics[width=0.06\linewidth]{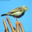} 
                    & \includegraphics[width=0.06\linewidth]{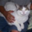} 
                    & \includegraphics[width=0.06\linewidth]{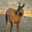}
                    & \includegraphics[width=0.06\linewidth]{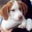}
                    & \includegraphics[width=0.06\linewidth]{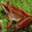} 
                    & \includegraphics[width=0.06\linewidth]{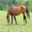}
                    & \includegraphics[width=0.06\linewidth]{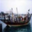} 
                    & \includegraphics[width=0.06\linewidth]{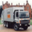} \\ 
                    &air-&auto-&bird&cat&deer&dog&frog&horse&ship&truck \\
                    &plane&mobile&&&&&&&& \\
     \midrule
     0 & \textbf{0.405} &                & 0.002          &                &                &   &   &   &   & \\
     1 &                & \textbf{0.455} &                &                &                &   &   &   &   & \\
     2 & 0.007          &                & \textbf{0.562} &                &                &   &   &   &   & \\
     3 &                &                &                & \textbf{0.602} & 0.004          &   &   &   &   & \\
     4 &                &                &                &                & 0.083          &   &   &   &   & \\
     5 &                &                &                &                & 0.006          & \textbf{0.482} &   &   &   & \\
     6 &                &                &                &                &                &                & \textbf{0.527} &   &   & \\
     7 &                &                &                &                & \textbf{0.387} &                &                & \textbf{0.556} &  & \\
     8 &                &                &                &                &                &                &                &  & \textbf{0.537} & \\
     9 &                &                &                &                &                &                &          &       & 0.004 & \textbf{0.471} \\
    10 & \textbf{0.583} &                & 0.002          &                &                &   &   &   &   & \\
    11 &                & \textbf{0.545} &                &                &                &   &   &   &   & \\
    12 & 0.005          &                & \textbf{0.434} &                &                &   &   &   &   & \\
    13 &                &                &                & \textbf{0.398} & 0.005          &   &   &   &   & \\
    14 &                &                &                &                & 0.056          &   &   &   &   & \\
    15 &                &                &                &                & 0.006          & \textbf{0.518} &   &   &   & \\
    16 &                &                &                &                &                &                & \textbf{0.473} &   &   & \\
    17 &                &                &                &                & \textbf{0.453} &                &          & \textbf{0.444} &  & \\
    18 &                &                &                &                &                &                &          &       & \textbf{0.455} & \\
    19 &                &                &                &                &                &                &          &       & 0.004 & \textbf{0.529} \\
    \midrule
    &  \multicolumn{10}{c}{Adversarial images}\\
    Labels            & \includegraphics[width=0.06\linewidth]{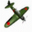} 
                    & \includegraphics[width=0.06\linewidth]{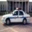} 
                    & \includegraphics[width=0.06\linewidth]{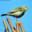} 
                    & \includegraphics[width=0.06\linewidth]{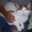} 
                    & \includegraphics[width=0.06\linewidth]{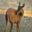}
                    & \includegraphics[width=0.06\linewidth]{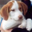}
                    & \includegraphics[width=0.06\linewidth]{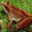} 
                    & \includegraphics[width=0.06\linewidth]{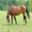}
                    & \includegraphics[width=0.06\linewidth]{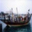} 
                    & \includegraphics[width=0.06\linewidth]{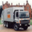} \\
                    &air-&auto-&bird&cat&deer&dog&frog&horse&ship&truck \\
                    &plane&mobile&&&&&&&& \\
     0 & \textbf{0.491} &                & 0.005          &   &   &   &   &   &   & \\
     1 &                & \textbf{0.544} &                &   &   &   &   &   &   & \\
     2 & 0.012          &                & \textbf{0.492} &   &   &   &   &   &   & \\
     3 &                &                &                & \textbf{0.470} & 0.004          &   &   &   &   & \\
     4 &                &                &                &                & 0.078          &   &   &   &   & \\
     5 &                &                &                &                & 0.007          & \textbf{0.511} &   &   &   & \\
     6 &                &                &                &                &                &                & \textbf{0.478} &   &   & \\
     7 &                &                &                &                & \textbf{0.435} &                &       & \textbf{0.490} &  & \\
     8 &                &                &                &                &                &                &       &       & \textbf{0.457} & \\
     9 &                &                &                &                &                &                &       &       & 0.002 & \textbf{0.503} \\
   10  & \textbf{0.486} &                & 0.004          &                &                &   &   &   &   & \\
   11  &                & \textbf{0.456} &                &                &                &   &   &   &   & \\
   12  & 0.009          &                & \textbf{0.499} &                &                &   &   &   &   & \\
   13  &                &                &                & \textbf{0.528} & 0.005          &   &   &   &   & \\
   14  &                &                &                &                & 0.054          &   &   &   &   & \\
   15  &                &                &                &                & 0.007          & \textbf{0.488} &   &   &   & \\
   16  &                &                &                &                &                &                & \textbf{0.522} &   &   & \\
   17  &                &                &                &                & \textbf{0.408} &                &       & \textbf{0.510} &  & \\
   18  &                &                &                &                &                &                &       &       & \textbf{0.539} & \\
   19  &                &                &                &                &                &                &       &       & 0.002 & \textbf{0.497} \\

    \bottomrule
  \end{tabular}
\end{table}

\begin{table}[t]
  \caption{Target~Training performs slightly better than Adversarial Training against attacks that do not minimize perturbation, both utilizing adversarial samples in training. We also compare with unsecured classifier performance. Further results in Appendix~C,~Table~\ref{target-with-adv-table-other-attacks} show that both Target~Training and Adversarial~Training provide defense against some attacks they have not been trained for, but not all.}
  \label{target-with-adv-table}
  \centering
  \begin{tabular}{llrrrrrr}
    \toprule
                               &           & \multicolumn{3}{c}{CIFAR10 (84.3\%)}  & \multicolumn{3}{c}{MNIST (99.1\%)} \\
                                                                 \cmidrule(r){3-5}   \cmidrule(r){6-8}   
    Adv. samples               &                  & Target & Adv.   & Unsec-   & Target & Adv.   & Unsec-   \\
    in training                & Adv. samples     & Train- & Train- & ured     & Train- & Train- & ured     \\
    (TT and AT)                & in testing       & ing    & ing    & Classif. & ing    & ing    & Classif. \\
    \midrule
    CW-$L_2$($\kappa=40$) & CW-$L_2$($\kappa=40$) & 77.7\% & 77.4\% &  8.5\%   & 98.0\% & 98.0\% &  0.7\%   \\
     PGD                  & PGD                   & 76.3\% & 76.2\% & 32.7\%   & 92.3\% & 91.7\% & 79.7\%   \\
    FGSM($\epsilon=0.3$)  & FGSM($\epsilon=0.3$)  & 72.1\% & 71.8\% & 11.8\%   & 98.0\% & 98.4\% & 10.0\%   \\
    \bottomrule
  \end{tabular}
\end{table}


\begin{table}[t]
  \caption{Expanded comparison of Target Training and Adversarial Training against attacks that do not minimize perturbation. Here, we show also performance against attacks, the adversarial samples of which have not been used in training. Both Target Training and Adversarial Training defend against some attacks that they have not been trained for, but not all. We also compare with unsecured classifier performance.}
  \label{target-with-adv-table-other-attacks}
  \centering
  \begin{tabular}{ll|rrr|rrr}
    \toprule
                     &            & \multicolumn{3}{c}{CIFAR10 (84.3\%)} & \multicolumn{3}{c}{MNIST (99.1\%)} \\
                                                       \cmidrule(r){3-5}   \cmidrule(r){6-8}   
    Adv. samples     &                        & Target  & Advers.& No     & Target  & Advers.& No     \\
    in training      & Adv. samples           & Train- & Train- & Defe-  & Train- & Train- & Defe-  \\
    (TT and AT)      & in testing             & ing    & ing    & nse    & ing    & ing    & nse    \\
    \midrule
                     & CW-$L_2$ ($\kappa=40$) & 77.7\% & 77.4\% &  8.5\% & 98.0\% & 98.0\% &  0.7\% \\
                     & CW-$L_2$ ($\kappa=0$)  & 71.3\% & 12.3\% &  8.8\% & 97.4\% &  1.5\% &  8.8\% \\
    CW-$L_2$       & DeepFool                 & 75.8\% & 13.2\% &  9.2\% & 97.6\% &  1.6\% &  1.3\% \\
    (conf=$40$)      & PGD                    & 10.0\% & 10.0\% & 32.7\% & 23.3\% &  2.9\% & 79.7\% \\
                     & FGSM($\epsilon=0.3$)   & 10.6\% &  9.9\% & 11.8\% & 56.6\% & 15.8\% & 10.0\% \\
                     & FGSM($\epsilon=0.01$)  & 48.9\% & 36.4\% & 40.4\% & 97.7\% & 97.8\% & 98.6\% \\
    \midrule
                     & PGD                    & 76.3\% & 76.2\% & 32.7\% & 92.3\% & 91.7\% & 79.7\% \\
                     & CW-$L_2$ ($\kappa=40$) &  7.3\% & 57.3\% &  8.5\% & 83.2\% & 98.4\% &  0.7\% \\
    PGD              & CW-$L_2$ ($\kappa=0$)  & 12.8\% & 12.7\% &  8.8\% & 94.3\% & 22.7\% &  8.8\% \\
                     & DeepFool               & 15.0\% & 13.0\% &  9.2\% & 86.5\% &  4.7\% &  1.3\% \\
                     & FGSM($\epsilon=0.3$)   & 10.7\% & 10.2\% & 11.8\% & 79.9\% & 95.4\% & 10.0\% \\
                     & FGSM($\epsilon=0.01$)  & 39.8\% & 41.5\% & 40.4\% & 98.2\% & 98.4\% & 98.6\% \\
    \midrule
                     & FGSM($\epsilon=0.3$)   & 72.1\% & 71.8\% & 11.8\% & 98.0\% & 98.4\% & 10.0\% \\
                     & FGSM($\epsilon=0.01$)  & 40.8\% & 42.1\% & 40.4\% & 98.5\% & 98.5\% & 98.6\% \\ 
    FGSM             & CW-$L_2$ ($\kappa=40$) & 49.9\% & 74.2\% &  8.5\% & 58.8\% &  1.1\% &  0.7\% \\
    ($\epsilon=0.3$) & CW-$L_2$ ($\kappa=0$)  & 12.5\% & 12.7\% &  8.8\% & 51.8\% &  1.1\% &  8.8\% \\
                     & DeepFool               & 12.7\% & 12.8\% &  9.2\% & 48.3\% &  1.2\% &  1.3\% \\
                     & PGD                    & 17.2\% &  1.2\% & 32.7\% & 72.6\% & 42.5\% & 79.7\% \\
    \bottomrule
  \end{tabular}
\end{table}

\end{document}